# Convolutional Neural Network Based Partial Face Detection

Md. Towfiqul Islam
*Department of Computer Science and Engineering*
*Daffodil International University*
Dhaka, Bangladesh
towfiqul15-11139@diu.edu.bd

Tanzim Ahmed
*Department of Computer Science and Engineering*
*Daffodil International University*
Dhaka, Bangladesh
tanzim15-10801@diu.edu.bd

A.B.M. Raihanur Rashid
*Department of Computer Science and Engineering*
*Daffodil International University*
Dhaka, Bangladesh
raihanur15-11069@diu.edu.bd

Taminul Islam*
*Department of Computer Science and Engineering*
*Daffodil International University*
Dhaka, Bangladesh
taminul@ieee.org

Md. Sadekur Rahman
*Department of Computer Science and Engineering*
*Daffodil International University*
Dhaka, Bangladesh
sadekur.cse@daffodilvarsity.edu.bd

Md. Tarek Habib
*Department of Computer Science and Engineering*
*Daffodil International University*
Dhaka, Bangladesh
tarek.cse@diu.edu.bd

*Abstract*— Due to the massive explanation of artificial intelligence, machine learning technology is being used in various areas of our day-to-day life. In the world, there are a lot of scenarios where a simple crime can be prevented before it may even happen or find the person responsible for it. A face is one distinctive feature that we have and can differentiate easily among many other species. But not just different species, it also plays a significant role in determining someone from the same species as us, humans. Regarding this critical feature, a single problem occurs most often nowadays. When the camera is pointed, it cannot detect a person's face, and it becomes a poor image. On the other hand, where there was a robbery and a security camera installed, the robber's identity is almost indistinguishable due to the low-quality camera. But just making an excellent algorithm to work and detecting a face reduces the cost of hardware, and it doesn't cost that much to focus on that area. Facial recognition, widget control, and such can be done by detecting the face correctly. This study aims to create and enhance a machine learning model that correctly recognizes faces. Total 627 Data have been collected from different Bangladeshi people's faces on four angels. In this work, CNN, Harr Cascade, Cascaded CNN, Deep CNN & MTCNN are these five machine learning approaches implemented to get the best accuracy of our dataset. After creating and running the model, Multi-Task Convolutional Neural Network (MTCNN) achieved 96.2% best model accuracy with training data rather than other machine learning models.

## I. INTRODUCTION

Face detection technology have experienced for the long time. Since then, it has been implemented in various ways in diverse fields [1]. It can be employed in a variety of industries, including entertainment, law enforcement and biometrics. Artificial neural networks (ANN's) have evolved from simple computer vision techniques to more complex models throughout the years. When it comes to facial recognition, it plays an important role. In the future, it will be able to identify faces in photographs and utilize that information for other reasons. AI-based computer technology is used to recognize human faces in digital photographs using face detection. Facial recognition is often the initial stage and has a considerable influence on the performance of subsequent operations within the project, especially in large scale projects. Helps identify which sections of the image or video are needed for a faceprint [2].

To extract human faces from bigger photos, face recognition technology makes use of machine learning and algorithms. These photographs generally contain many non-facial elements in the background, such as houses, landscapes, and other body parts. The use of machine learning techniques can make it easier to spot a face among different objects. Face detection techniques employ a variety of machine learning and data mining algorithms to locate and collect interesting data from the internet and local. An interesting example of data mining is image mining [3], which involves using machine learning that explores information, images' data dependence, and unambiguous patterns. Faces are usually detected by looking at specific factors, such as the geometrical outlay of a face or the positions of eyes, nose, ears, and mouth. For face detecting, ANN [4] can be used. It actually can be derived from object detection technology such as Girschik et al. [5] showed in their research.

We discovered a problem where a face is covered or a part of the face is covered. And, for this reason, most computer vision techniques cannot find faces even when there are visible points of faces and human eyes can recognize them whereas the machine cannot. Since a face is the most important part of a human to recognize the person with, it takes a lot of priority to detect it.

Sometimes many people face a scenario that we have an image or video that is not of high quality and the person in it has done something that needs to be traced back to the person but using general approach face detection techniques most of the data about that person are lost and it is sometimes hard to detect a face, and it becomes harder for people to process any further for more analysis [6]. That's why this work is supposed to build a model by using a machine learning algorithm that will be able to detect the face of the person and it can be further processed in different tasks.

In this research work a number of machine learning techniques have used to complete this work, including the Convolutional Neural Network (CNN), the Cascaded Convolutional Neural Network, Haar Cascade Classifier, and the Deep Convolutional Neural Network (DCNN) and Multi-Task Cascaded Convolutional Networks (MTCNN) were presented and used. The MTCNN model proved to be the most accurate & provides the best accuracy.

This paper has an extra five parts. The "Related Work" section includes a background analysis. Methodology has been clearly detailed in this section. In order to demonstrate the proposed idea, graphs and tables are employed. In the Result section, all of the performances are thoroughly analyzed. Discussion and Conclusion portions of this study are aimed to assist in making a choice. This part also serves as a wrap-up and discussion on where the research is headed.

## II. RELATED WORKS

Zhang et al. [7] conducted research that included fresh 32,203 photos that were utilized in three subgroups of testing. Hard sample mining was also suggested, which would automatically boost performance without the need for manual sample sections. To begin with they developed brand-new tools for detecting faces that achieved superior accuracy over the FDDB and wider face benchmarks. AFLW benchmark face alignment accuracy was surpassed, but real-time performance was maintained.

In this study, Li et al. [8] employed cascaded CNNs to recognize faces, they neglected the natural link among facial features localization and regression problems in favor of relying on face detection's bounding box calibration. This project will continue to improve in terms of accuracy and output using a different machine learning technique. It's important to note that each sort of detection method has various strengths and drawbacks. The paper's limitation is that they cannot achieve more precise results without going through the process of systematic analysis. Chen et al. [9] collaborated on the alignment and detection of pixel value difference features using a random forest based on pixel value difference features.

Multiple CNNs are used by Zhang et al. [10], but the performance of multi-view face identification is still restricted by the weak face detector's detection windows. The detector's power was boosted by mining hard samples throughout the training phase. Hard sample mining is traditionally done offline, which increases the amount of time it takes to process a sample. Using unified cascaded CNNs, they came up with a novel framework for merging these two objectives. There are three steps to their suggested CNNs. P-Net, a shallow CNN, was used in the first step to quickly generate candidate windows. Then, using a more complicated CNN known as R-Net, it simplifies the windows to exclude a huge number of non-face windows. Finally, O-Net employs a third very powerful CNN to improve the precision and output face landmark locations [11]. The multiple CNNs designed have performance limitations where some Filters may not be able to create discriminative descriptions because they lack a wide range of weights. Besides, to put it another way, face recognition and classification is a difficult binary classification job. As a result, fewer filters may be required, but they must be more discriminating. There have used CNN, Cascaded CNN, Random Forest, and even multi-task CNN in multi-view face detection. But they got the best accuracy of 95.4% in multi-task CNN using the proposed model.

Face detection is no different in that finding labeled datasets is always a difficulty for machine learning researchers. Using different methods to collect data and produce models they have come up and developed a unique technique. To come up with the final dataset they used a new data collecting method. They also used different datasets that were used in many pieces of research to come up with their perfect model. There was no statistical analysis done to see if the difference between CCNN and DCNN in terms of performance was significant since they had a huge dataset.

## III. METHODOLOGY

This section provides an overview of the study's primary approach. Fig 1 depicts the general flow of this research. This section focuses on the dataset's origins and characteristics. In addition, the surrounding background is discussed. In the final part of this chapter, certain categorization models and assessment processes are briefly discussed.

The dataset has been created from the localization of our researchers. Faces can be analyzed in four ways: full, full with a mask, right or left side of the face. Using preprocessing techniques, data that is noisy and inconsistent can be cleaned up. The quality of the final product has been improved by a variety of preprocessing methods. We also had to remove faulty data in order for the model to function correctly. Preprocessed attributes from the face dataset are used to build a feature set for the classification model. Once this is done, several Classification algorithms are trained to be used in trials linked with our research. Fig. 1 shows the proposed model workflow of this research work.

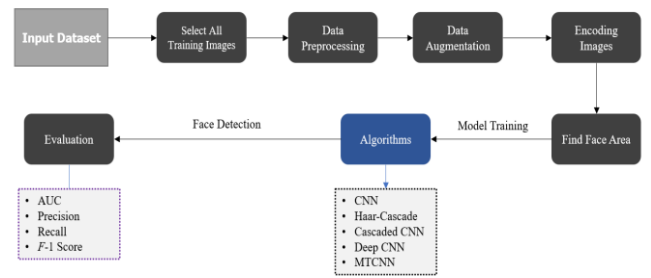

Fig. 1. Proposed model workflow

### A. Data Description

Data collecting is typically a difficult aspect of a research study, and this is no exception. To gather all of the new information in a short and hard period of time from the field was a difficult task. This work has gathered a total of 627 data from Bangladeshi peoples in 4 categories. Data was collected from 4 angles of a person's face. Front with mask, Front without mask, Right side with mask, Left side with mask. All of these data have collected manually & this work have created a fresh dataset. Table 1 states the amount of collected data.

TABLE 1: STATISTICS OF COLLECTED DATA

| Category | Data Size |
|---|---|
| Front Face | 159 |
| Front Face with Mask | 157 |
| Face Left Side | 155 |
| Face Right Side | 155 |
| Total | 627 |

Some of collected data samples are shown below:

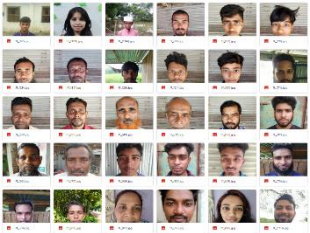
Fig. 2. Front face without mask

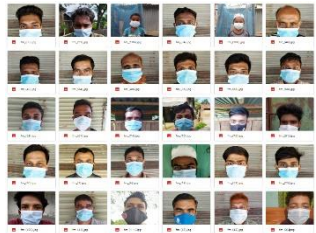
Fig. 3. Front face with mask

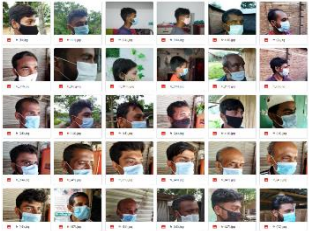
Fig. 4. Right side face with mask

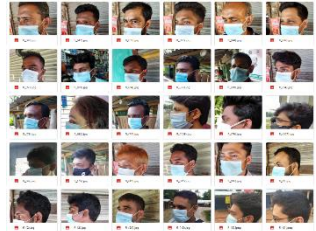
Fig. 5. Left side face with mask

Fig 2-5 represents the examples of collected data. This study gathered these 4 category images for classification & further pre-processing.

*B. Data Pre-Processing*

The first step in conducting research is preprocessing data. To begin, we must process the data we've collected. There are several sources we obtain our data for this reason. We start by preparing the data to fix this. These data sets are split down into a wide range of numerical values. This data is processed one at a time. Machine learning and deep learning models cannot be applied to data that is only in the form of images, it can handle numerical data only.

*1) Reduce of Dimension:* Unnecessary qualities are responsible for increasing the length of time it takes to complete an operation. Only the facial landmarks take priority here in this model.

*2) Noise Removal:* All data must be clear and free of noise. Otherwise, there might be some unanticipated issues that need to be dealt with. Corrupted and unclear images are harmful to the model. The dataset is improved by removing or cropping the images.

*3) Pixel brightness transformations:* The brightness of a pixel can be altered by performing a brightness transformation, which is dependent on the pixel's own attributes. If the input pixel's color is red, then the output's color is red. Color correction and transformations are just two examples of these types of operators.

*C. Proposed Model Working Procedure*

Data has been collected from four different angles. The main goal was to find the best machine learning model that can detect faces better. In this work, five machine learning algorithms have applied to find the best accuracy on this fresh dataset. These are Convolutional Neural Network, Cascaded CNN, Haar-Cascade, Deep CNN, and Multi-task Cascaded Convolutional Network. Initially, the data were split into two groups. 80% of the data have utilized for training, and 20% for testing. And this is also what we expect to observe in this model.

*1) Feature Selection and Extraction:* The retrieved characteristics are classified using CNN, Cascaded CNN, Deep CNN, and Haar-Cascade. Python is used to implement all of the classifiers. Machine learning libraries in python are extensive and feature-rich. There are two popular machine learning software programs called 'SKLEARN' and 'PANDAS'. Precision, recall, F1-measure, and accuracy are used to evaluate the classification model's performance. Besides, the Keras library was used to create the MTCNN model, which was then tested. It is possible to make a model using a complete FaceNet. The precision, recall, F1-measure, and accuracy are used to evaluate this classification model's performance.

*2) Machine Learning Models:* Five machine learning algorithms have applied to find the best accuracy on this fresh dataset. These are CNN, Cascaded CNN, Haar-Cascade, Deep CNN, and MTCNN. This section has placed a strong emphasis on these models.

*a) Convolutional Neural Network (CNN)*

A CNN is a type of deep neural network that is most typically used to assess visual data in the context of deep learning. Instead of employing a neural network, it makes use of a technique known as Convolution. There are three planes to a matrix of pixel values in an RGB image, but there is only one plane in a grayscale image. To obtain the convolved feature, a kernel (3×3 matrix) is applied to the input image. Then the next layer receives the convolved feature. It is a mathematical function that computes a weighted sum of many inputs and produces an activation value in CNNs, which are made of several artificial neurons [12].

*b) Cascaded CNN*

If you've ever seen a decrease in performance when increasing the Intercept over Union (IoU) threshold, Cascaded R-CNN, commonly known as Cascaded CNN, is an object detection design that addresses this issue. More detector stages in the cascade are successively more discriminating against near false positives, making it a multi-stage variant of the R-CNN. The R-CNN stages are trained in a cascade, one after the other, successively [13].

*c) Deep CNN*

Deep learning is a technique used in AI to create intelligent machines. Artificial neural networks are the inspiration for this system (ANN). An ANN is a neural network that uses numerous layers of neurons to perform complicated analysis on enormous quantities of data. To discover patterns in photos and videos, Deep CNN are the most prevalent [14].

*d) Haar-cascade Classifier with boosted Gaussian Feature*

Haar feature-based cascade classifiers are object detection methods. This technique necessitates a large number of photos, both positive and negative, that contain and do not contain faces. Then, using a convolutional kernel, it extracts characteristics from it. The combination of pixels under the white rectangle is subtracted from the total of pixels under the black rectangle to get a single feature value [15]. Using Adaboost, the enormous number of options in a (n × n) window is narrowed down to only the best ones. The anisotropic Gaussian feature is the Gaussian feature employed in this case. To build the weak classifiers, a fresh batch of local filters are needed.

*e) MTCNN*

"Multi-task Cascaded Convolutional Network" or MTCNN is a deeper cascaded multi-task architecture that uses the intrinsic connection between the deep learning methodologies to increase their performance. There are three levels of convolutional networks in this architecture that predict facial landmark locations from coarse-to-fine resolution. Which is the hyperparameter optimization of a neural network during trying out different combinations of the hyperparameters and evaluating the performance of the network [16]. Due to a large number of parameters and the big range of their values, it discretizes the available value range into a "coarse" grid of values to estimate the effect of the value of the parameter and then performs a "finer" search around it to optimize even further. There are three convolutional networks in MTCNN, that are P-Net, R-Net, and O-Net, respectively. Possibility frames and their subset regression vectors are obtained using P-Net, also known as Proposal Net, a fully convolutional network. First from top left corner, it scans the picture from (0,0) to (12,12) and scales a 12×12 matrix kernel that runs through the image scanning for faces. The weights and biases of the P-Net have been trained to provide a pretty accurate boundary for each 12×12 kernel. The R-Net, also known as the Refine Net, performs calibration with box regression and NMS, or Non-Maximum Suppression, which is a method that minimizes the number of bounding boxes to be used for merging candidates. It resizes the bounding box arrays, to 24×24 pixels and normalizes the values between -1 and 1, after padding. Lastly, the O-Net which is slightly different than the both previous P-Net and R-Net. The bounding box coordinates, the five facial landmarks coordinates, and the level of confidence for each box are the three outputs. We then run it through the NMS one final time, this time eliminating the lowest confidence level boxes while also standardizing coordinates for the bounding box and face landmark. Only one bounding box for each face is allowed in Fig. 6 proposed by [17].

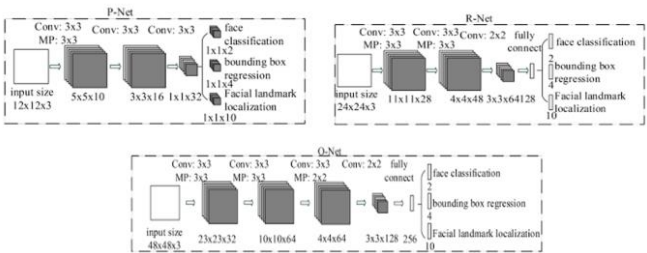

Fig. 6. MTCNN neural networks [17]

*D. Performance Measurement Unit*

Different authors used a variety of measures to assess the success of their models. Despite the fact that the majority of the studies employed numerous indicators to assess their effectiveness, a small number also used a single metric. In this work accuracy, precision, recall & $F_1$-Score is considered for evaluating this research work. These four-measurement unit is the best for face data analysis.

*1) Accuracy*

The percentage of successfully predicted items to all feasible predictions is what constitutes a facemask detection model's accuracy. The precision is defined in the below equation.

$$\text{Accuracy} = \frac{TP + TN}{TP + FP + TN + FN}$$

*2) Precision*

Probability of forecasting a mask bounding box that fits the ground truth is called precision. This is a count of how many masks the model successfully identifies out of the total number of masks in the image. It is the ratio specified in the below equation between TP and all positives.

$$\text{Precision} = \frac{TP}{TP + FP}$$

*3) Recall*

A detector's recall refers to its capacity to accurately locate and identify all conceivable facemasks or actual values. In the below equation defines it as the ratio between TP and the total of TP and FN.

$$\text{Recall} = \frac{TP}{TP + FN}$$

*4) $F_1$-Score*

Due to its reliance on both precision and recall, this is referred to as the harmonic mean. Below equation is an expression of a mathematical equation for memory retrieval.

$$F_1 - \text{Score} = 2\left(\frac{\text{Precision} \times \text{Recall}}{\text{Precision} + \text{Recall}}\right)$$

## IV. EXPERIMENTAL EVALUATION

Total five machine learning algorithms have applied to this fresh dataset. There is a tight comparison between one to other algorithms. MTCNN achieved the best accuracy, precision, recall, and $F_1$-Score in between five algorithms. MTCNN achieved the best 96.2% accuracy where Cascaded CNN & Deep CNN achieved 71% and 70% accuracy respectively. We found 68% accuracy in CNN and 65% accuracy on the Haar-Cascade classifier boosted with Gaussian Features. Below Table 2 illustrates the result comparison between the five machine learning algorithms.

TABLE 2: RESULT COMPARISON BETWEEN FIVE MACHINE LEARNING ALGORITHMS

| Algorithms | AUC | Precision | Recall | $F_1$-Score |
|---|---|---|---|---|
| Haar-Cascade | 0.652 | 0.615 | 0.635 | 0.625 |
| CNN | 0.684 | 0.643 | 0.662 | 0.652 |
| Deep CNN | 0.70 | 0.687 | 0.691 | 0.69 |
| Cascaded CNN | 0.713 | 0.682 | 0.702 | 0.692 |
| MTCNN | 0.962 | 0.89 | 0.78 | 0.823 |

As illustrated in Fig. 7. the accuracy of MTCNN is higher than that of the other algorithms when compared to their combined accuracy.

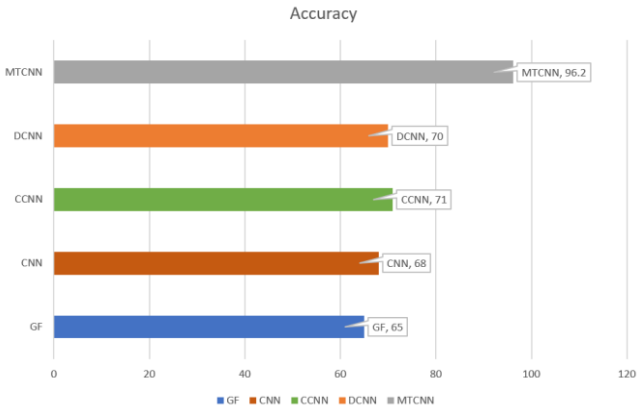

Fig. 7. Comparative representation of highest accuracy

The proposed model's output identifies the faces presented in images. The face detected is either true, false, or partially false. The research found the best performance in the MTCNN method relative to all other algorithms. Table 3 summarizes the performance report of MTCNN.

TABLE 3: PERFORMANCE OF MTCNN

| Algorithm | Accuracy | Precision | Recall | $F$1-Score |
|---|---|---|---|---|
| MTCNN | 92.6% | 0.89 | 0.78 | 82.3% |

Samples of this model's detected faces are given below. In these images, it is clearly seen that this model has detected the face without a mask with 92.83% of accuracy shown in Table 4. However, as it stands the accuracy for partial face detection is still lower due to the lack of sufficient training data. It can detect the face with an accuracy of 79.703%. The landmarks of a face are detected here which are eyes, nose, and two corners of the lips are visible here in the image without a mask. But in the image with a mask on only shows two facial landmarks which are the eyes. So, it can be said that the face detected with this accuracy with this little dataset is satisfactory. But the total accuracy takes a hit when it comes to the side of a face. The accuracy doesn't even reach 50%. So, it can be said that it still has a lot of ways ahead for improvement.

TABLE 4: DETECTED FACED WITH WITHOUT MASK FROM DIFFERENT ANGLE

| Sample Picture | Class | Accuracy |
|---|---|---|
|  | Front face detection with mask | 79.703% |
|  | Front face detection without mask | 92.83% |
|  | Left face detection with mask | 45.4% |
|  | Right face detection with mask | 41% |

## V. DISCUSSION

Partial face detection is the focus of most of the current investigation work in these fields. The purpose of this research was to determine whether or not machine learning algorithms can improve and detect partial faces with the most accuracy they can. The classification method used in this study performed well. Our thesis has revealed that most of the devices only use basic face detection technologies due to their heavy implementation methods. That's why we tried to make it as simple as possible to run even in a device of low specification.

TABLE 5: SUMMARY WITH OTHER PUBLISHED WORK

| Ref | Yr | Contribution | Dataset | Models | Accuracy |
|---|---|---|---|---|---|
| This work | 2022 | Developed partial face detection and alignment using MTCNN | Fresh data | MTCNN | 96.2% |
| [18] | 2018 | Developed joint face detection and alignment using MTCNN | FDDB [19], WIDER FACE [20], AFLW [21] | MTCNN | 95.4% |
| [22] | 2014 | Developed joint cascaded face detection and alignment | FDDB [23], AFW, CMU-MIT [24] | Cascaded CNN | 95.00% |
| [25] | 2014 | Developed Multiview face detection | FDDB [26] | Deep CNN | 85.00% |
| [27] | 2017 | Developed CNN for robust face detection | Yale [28], ORL [29] | CNN | 92.27% |
| [30] | 2007 | Developed boosted face detection using Gaussian Feature | XM2VTS [31], FERET [32] | Haar-cascade Classifier | 94.00% |

This research focuses on identifying faces through the use of well-known machine learning techniques. Our researchers collected data locally from person to person. Following feature extraction and model construction, MTCNN attained the highest accuracy of 96.2% that performs better than other related works shown in the Table 6. Table 5 clearly states the comparison between previous work & this work. The model was tested using newly collected data. Collecting high-quality data can help enhance accuracy. In the future, we hope to test our proposed technique on a bigger, more varied dataset to get over our existing limitations. This model can correctly identify faces as well as partial faces which will be an improvement for all. It was a challenging task to collect all quality fresh data in this pandemic situation, The result satisfies us in terms of the new dataset. If we work hard to collect the best possible quality data, then this model should achieve more accuracy in every sector.

## VI. CONCLUSIONS

Researchers have been working on Face detection for a long time. Face detection is necessary in many sectors of our current world. Technological companies that try to make the best of their devices or services are also trying to get into the field of face detection for its' extended uses. New AI software and services are being developed to improve the user experience. However, face detection still has a lot of ways to improve. The more improved it will be the better implementation of it will be available. It has yet to become a lightweight package that everyone can use on any low-end device. This work has gathered total 627 fresh images from Bangladeshi peoples on four different angels & applied five machine learning algorithms to detect the face in terms of angles & mask. All the algorithms perform good but MTCNN performs better than all other algorithms. This work has achieved the best result from Multi-task Cascaded Convolutional Network (MTCNN) with 96.2% accuracy. This work has revealed that most of the devices only use basic face detection technologies due to their heavy implementation methods. That's why we tried to make it as simple as possible to run even in a device of low specification. Collecting all quality new data in this pandemic circumstance was a difficult endeavor, but the outcome satisfies us in terms of the new dataset. If we work diligently to obtain the highest-quality data possible, our model should attain greater accuracy in every field.